\documentclass[11pt]{article}
\usepackage[preprint]{neurips_2026}

\usepackage[utf8]{inputenc}
\usepackage[T1]{fontenc}
\usepackage{url}
\usepackage{hyperref}
\usepackage{xcolor}
\usepackage{amsmath,amssymb,amsfonts}
\usepackage{graphicx}
\usepackage{comment}
\usepackage{subcaption}
\usepackage{authblk, graphicx, tikz}

\title{Texture Representations in Deep Vision Models: Comparing CNNs, Vision Transformers, and Human Perception
}

\author{
Ludovica de Paolis\textsuperscript{1,}\thanks{Correspondence to Ludovica de Paolis: \href{mailto:ldepaoli@sissa.it}{ldepaoli@sissa.it}} ,
Marco Baroni\textsuperscript{2,3},
Alessandro Laio\textsuperscript{4},
Eugenio Piasini\textsuperscript{1}
\\
\textsuperscript{1}Department of Neuroscience, International School for Advanced Studies (SISSA), Trieste, Italy\\
\textsuperscript{2}Department of Language and Translation Sciences, Pompeu Fabra University, Barcelona, Spain\\
\textsuperscript{3}ICREA, Barcelona, Spain\\
\textsuperscript{4}Department of Data Science, International School for Advanced Studies (SISSA), Trieste, Italy\\
}

\date{}

\begin{document}
\maketitle

\begin{abstract}
In computational vision science, Convolutional Neural Networks (CNNs) have emerged as a popular model of biological vision because of the alignment they can exhibit with neural and behavioral data in humans and animals. However, it remains unclear to what extent this alignment persists for visual tasks that extend beyond the canonical object-recognition paradigm based on well-defined semantic content.
In this study, we diverge from the common object-centric view by focusing on another aspect of vision: texture perception. We consider textures of different complexity generated with three different algorithms from the same source images. Using a rank-based statistic, we quantify the information encoded in the internal representations of a CNN and three Vision Transformers (ViTs), and we compare the similarity  of these representations to those inferred from human psychophysics data. We find that the representation of textures is aligned in different ViTs, but not between the ViTs and the CNN; that ViTs form similar representations for textures of different complexity; that human performance in recognizing textures can be better predicted from ViTs representations rather than CNN representations. Taken together, these results suggest that ViTs may capture more faithfully than CNNs how texture patterns are visually processed by humans, and that the representations of texture stimuli in computational models may be driven by the network architecture. 
\end{abstract}

\section{Introduction}

Convolutional neural networks (CNNs) are ubiquitous in computer vision and in vision neuroscience, where they are commonly employed as models of the ventral visual stream and of the biological mechanisms underlying object recognition~\citep{Lindsay2021,Kriegeskorte2015DeepNeuralNetworks}). A pivotal moment in the history of CNNs was the release of AlexNet~\citep{KrizhevskySutskeverHinton2012}, which demonstrated significantly higher accuracy than other vision models in an object classification task connected to the ImageNet database~\citep{deng2009imagenet}. This success reinforced the centrality of the object recognition paradigm in machine learning and computational vision neuroscience, which led to numerous advances in the following years. However, in real life scenarios image perception and representation is much more diversified than just recognizing objects. Consequently, parallel research directions in vision science have been focusing on aspects of vision that cannot be addressed within the framework of object recognition. One of these tasks is texture perception: visual textures are an acknowledged method to study complex real world vision phenomena \citep{victor2017}. 
An early formal definition of textures was provided by Julesz~\citep{julesz1962, julesz1981},  stating that textures can be seen as families of patterns that share certain local regularities. Julesz also conjectured that textures could be modeled via low-order statistics \citep{julesz1973}. This conjecture became very influential in the development of computational models of textures, which later resulted in the release of texture synthesis algorithms that operationalized the conjecture by means of low-level correlations. The most prominent models are those by Portilla \& Simoncelli \citep{portilla2000}, Victor \& Conte \citep{victor2012}, and Gatys et al., \citep{gatys2015}. The model by Victor \& Conte synthesizes artificial binary textures by means of local correlations in \(2 \times 2\) pixel gliders in a 10-dimensional parametrized space. The model by Portilla \& Simoncelli produces gray-scale textures by means of a fixed set of statistics of the wavelet coefficients of a given input image. The algorithm by Gatys synthesizes naturalistic textures by computing local correlations among the feature maps extracted from a CNN pretrained on object recognition. Since CNNs are regarded as a good model for object recognition in human and primate vision~\citep{schrimpf2018, storrs2021, guclu2015, yamins2014, Lindsay2021}, the success of this approach indirectly linked texture perception mechanisms to those underlying object recognition. However, it was recently shown that the quality of the texture representation associated with different CNNs does not correlate with well-established metrics of representational and behavioral alignment of these same networks with biological vision~\citep{depaolis2026}. This finding suggests that common measures of alignment may focus too narrowly on object recognition, and may underweight other perceptual phenomena like texture perception. This, in turn, may imply that network architectures different from CNNs may be useful representational models when more diverse tasks other than object classification are taken into account.

In the last few years, Vision Transformers (ViT) have emerged as another extremely effective architecture for  computer vision tasks \citep{chen2020igpt}. ViT can perform object classification \citep{radford2021}, feature distillation \citep{oquab2024dinov2} and image generation \citep{chen2020igpt, Yu2022ImprovedVQGAN}. ViTs belong to the family of Transformers, which are deep neural network models equipped with the attention mechanism operationalized with multi-head attention layers, which allow to produce internal representations of the input images encoded in an array of tokens by attending all the possible combinations of positions between image patches of a fixed size. While CNNs produce image representation leveraging local correlations with convolutions and pooling operations, ViTs form image representations by means of both local and long range correlations among image patches. The attention mechanism effectively integrates both levels of correlations, which in turn produce image representations robust to spatially-bound information, allowing ViTs to form stable and consistent representations across data that preserve the same meaning or structure.



We are interested in  studying how CNNs and ViTs represent textures, in comparing their performance, and in comparing the models' performance against human subjects' perception. We perform a comparative analysis of CNNs, Vision Transformers, and human subjects using a large image dataset spanning a continuum of visual complexity. The dataset combines two existing stimulus databases, one containing images of objects and one containing images of textures. Starting from the texture database, we generate synthetic textures with three different algorithms, producing stimuli with distinct levels of complexity. We then extract texture representations from one CNN and three ViTs and compare them with human behavioral performance in a texture-perception task. The models' representations are analyzed using a rank-based statistic designed to quantify information in high-dimensional feature spaces. Human performance is measured with an odd-one-out paradigm, as in \cite{wallis2017,jagadeesh2021}, in which participants identify the odd stimulus among sets of similar textures with different levels of complexity.

We find that 
(i) the three ViTs align among themselves in representing textures of different types of complexity, but the CNN does not align with ViTs; (ii) human perception of textures of different types strongly aligns with the texture representations produced by the three ViTs but does not align with the texture representations produced by the CNN.

\section{Methods}
\subsection{Stimulus datasets}

\paragraph{Textures and objects databases.}
To create the dataset for our study we assemble two existing databases, one containing images of objects and one of textures. We use the Describable Textures Dataset (DTD) as a texture dataset and as a source to generate synthetic textures with the three algorithms by \cite{victor2012, portilla2000, gatys2015}. Using this procedure, we obtain a large pool of textures with distinct levels of complexity, where each datapoint is generated from one datapoint of the source, DTD. In this way, our natural and synthesized texture datasets are naturally aligned at the single-image level. This passage is key as the rank-based metric of our choice quantifies the variations in information across different subsets as changes in the local neigborhoods.

\begin{figure}[ht]
\centering
\begin{tikzpicture}[x=1cm,y=1cm,>=latex, line cap=round]
  \draw[->, thick] (0,0) -- (12,0);
  \node[above]       at (6.5,0) {\bfseries Textures complexity spectrum};
  \node[above left]  at (0,0)   {Simple};
  \node[above right] at (12,0)  {Rich};

  \def\imgw{1.25}   
  \def\gap{0.9}    
  \pgfmathsetmacro{\pitch}{\imgw+\gap} 
  \pgfmathsetmacro{\xstart}{\imgw/2}   
  \def\yoff{-1}  

  \foreach [count=\i] \file/\cap in {
    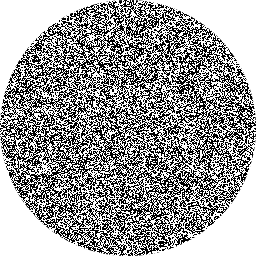/Noise (\scriptsize Independent),
    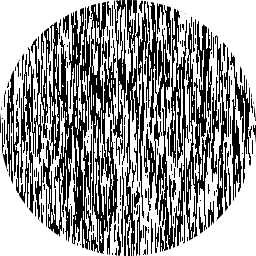/V\&C,
    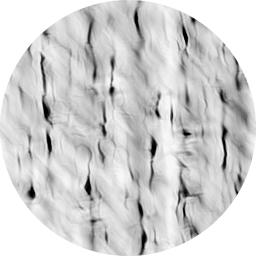/P\&S,
    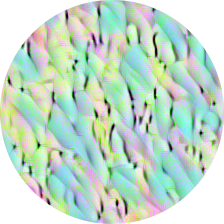/G,
    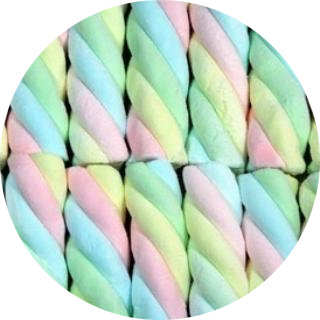/DTD (\scriptsize Source),
    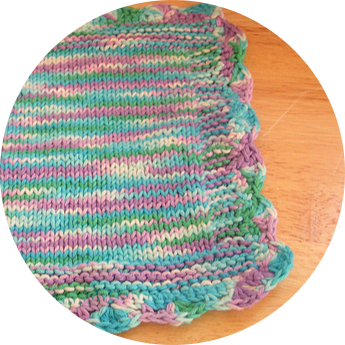/Object (\scriptsize Independent)
  }{
    \pgfmathsetmacro{\x}{\xstart + (\i-1)*\pitch}
    \node at (\x,\yoff) {\begin{tabular}{@{}c@{}}
      \includegraphics[width=\imgw cm]{\file}\\[-1mm]
      \footnotesize \cap
    \end{tabular}};
  }
\end{tikzpicture}
\caption{\textbf{Example stimuli arranged along a putative spectrum of texture complexity.} In the Noise sample, each pixel is sampled independently at random as black or white with equal probability. The stimuli labeled as V\&C, P\&S, and G are obtained by the DTD image classified as ``braided'' (marked as ``source'') and by application of the respective algorithm. The Object image was chosen by hand for illustrative purposes from the Object dataset.}
\label{fig:texture_spectrum}
\end{figure}

As shown in Figure ~\ref{fig:texture_spectrum}, the textures generated with the Victor \& Conte algorithm (\textbf{V\&C}) are simple, coarse, and show little structure; the textures by Portilla \& Simoncelli (\textbf{P\&S}) are intermediate, greyscale and preserve some global structure; the textures by the Gatys' algorithm (\textbf{G}) are more complex, show good local structure and color; finally, the \textbf{DTD} textures are the most perceptually and semantically complex. We expand the description of the synthetic textures below. DTD was developed by \cite{cimpoi2014} to have both naturalistic quality and human-like semantic accuracy. It consists of 5640 images of realistic textures collected from the internet, equally distributed in 47 semantic categories. The class labels are selected from the ``Texture Lexicon'', an independent study by \cite{bhushan} in which participants were asked to build a set of words describing categories of textures. The naturalistic status of its textures and the independent linguistic human agreement make DTD relevant for our scope, because the textures contain relevant semantic information.

For the objects database, we considered a subset of 5640 images from ImageNet-1K validation set \citep{russakovsky2015imagenet, deng2009imagenet} (henceforth \textbf{Object}). ImageNet is the most popular dataset for training neural networks in computer vision and the most used for studying object classification and perception with computational models. We therefore use it as a reference stimulus for representing complex objects with a well-defined semantic content provided by the class labels. Finally, we also consider 5640 images constituted by fully random black and white pixels (henceforth \textbf{Noise}) to exemplify images with neither semantic content nor lower-level systematic structure.

\paragraph{Synthetic texture stimuli.}\label{synth_text_stim}
As mentioned above, we use three different synthesis algorithms: \textbf{V\&C} by Victor \& Conte, \textbf{P\&S} by Portilla \& Simoncelli and \textbf{G} by Gatys, to generate synthetic stimuli that show different levels of image complexity. We use textures from DTD as an input for each algorithm as follows:

\begin{itemize}
\item \textbf{V\&C:} We generated V\&C type of textures using the implementation by \cite{PiasiniMetex} of Victor \& Conte's model. This method samples maximum-entropy textures with fixed correlation statistics among neighboring pixels. The algorithm provides an elegant parameterization of texture space, which is suitable for highly-controlled experiments, but it produces textures that are perceptually distant from those found in natural images: they feature black and white squares arranged in a regular pattern. These textures contain no semantic content, and constitute the lowest-level subset in our semantic content hierarchy. (Figure~\ref{fig:texture_spectrum}: V\&C).
\item \textbf{P\&S:} We generated P\&S type textures using the Python adaptation of the original Matlab implementation by Portilla \& Simoncelli, publicly available in the GitHub repository
\href{https://github.com/LabForComputationalVision/textureSynth}{textureSynth} by~\cite{portilla_simoncelli_texturesynth}. This algorithm computes the correlations among parameters of a wavelet decomposition, generating synthetic textures that statistically match the ones found in input images. The generated textures are in grayscale, preserve most of the structure in periodic images, and serve as image samples of mid-level complexity (Figure~\ref{fig:texture_spectrum}: P\&S).
\item \textbf{G algorithm:} We generated G type textures with the algorithm by \cite{gatys2015}, implemented by \cite{depaolis2026} publicly available in the GitHub repository
\href{https://github.com/ludovicadepaolis01/perceptual_misalignment}{perceptual\_misalignment}. The algorithm effectively defines a texture ensemble based on the Gram matrices formed by computing spatial correlations among intermediate layer activations of a CNN trained for object classification (VGG-19, \cite{simonyan2015vgg}). Texture synthesis is achieved by optimizing the pixels of an initially random image to match the statistical structure of the original texture via gradient descent. The generated textures preserve the colors and the perceptual content of the original image, as well as most of the semantic content, serving as image samples with middle-to-high complexity (Figure~\ref{fig:texture_spectrum}: G). Further details on the algorithms' implementation and data preprocessing are provided in Appendix ~\ref{stim_data_gen}. 
\end{itemize}

\subsection{Models}\label{models}
We extracted the features of all the stimulus datasets described above from all the layers of four vision neural networks, in order to understand how they represent different kinds of textures and run a comparative analysis. We tested three ViTs and one CNN. The models vary in architecture, training objective and task, but they share the training dataset: object images from ImageNet \citep{deng2009imagenet}. For the ViT family, we tested the visual encoder of \textbf{CLIP} \citep{radford2021}, a supervised ViT pretrained with a contrastive image-text objective; \textbf{DINO-v2} \citep{oquab2024dinov2}, a self-supervised image features extractor; \textbf{iGPT} \citep{chen2020igpt}, an unsupervised autoregressive decoder for image generation. For the CNN family we tested \textbf{VGG-19} \citep{simonyan2015vgg}, a supervised image classifier commonly used as a model of image processing in the ventral visual stream (\cite{GeirhosEtAl2017, schrimpf2018}). When performing this analysis, we controlled for parameter count, task, training objective, and architecture. VGG-19, CLIP, DINO-v2 number of parameters is around the same order of magnitude; VGG-19 and CLIP can perform object classification; all the models were trained with different objectives; but ViTs share the attention mechanism in their architecture, while VGG-19 has convolutions. A summary of the models' details is provided in Appendix ~\ref{mod_det}, Table~\ref{table:models_details}.

\subsection{Human psychophysics}
We want to obtain a measure of human agreement in forming relationships between different types of textures, and to compare their performance to that of CLIP, DINO-v2, iGPT, and VGG-19 respectively. We designed a psychophysics experiment to measure the perceptual distance between textures belonging to different subsets as described above, by assessing the similarity between texture images belonging to the same semantic class (``braided'') but displaying different complexity levels.  Inspired by the psychophysics literature on texture perception, we adopted an odd-one-out paradigm \citep{wallis2017, jagadeesh2021}, implemented as follows:
\begin{itemize}
    \item on each trial, three image stimuli are presented simultaneously, at three different locations that are equidistant from a fixation cross in the center of a computer screen;
    \item two of them (the ``reference'' stimuli) belong to the same subset, while the third (the ``odd'') to a different one (e.g.\ two DTD vs. one P\&S);
    \item the participants' task is to identify the odd stimulus;
    \item we only consider texture stimuli from V\&C, P\&S, G and DTD subsets;
    \item within the same trial, the three stimuli belong to the same DTD semantic class;
    \item stimuli are shown for 200ms, preventing visual search and engaging preattentive judgement.
    \item following a standard procedure~\citep{wallis2017}, we displayed all stimuli in grayscale regardless of their subset or class.
\end{itemize}

We collected data from 51 participants. Further details about the experimental setup, a visual display of the stimuli, and an illustration of the experimental flow can be found in Appendix ~\ref{hum_beh_exp}, Figures~\ref{fig:stimuli},~\ref{fig:flow}. 

\subsection{Information Imbalance}
We use Information Imbalance (\textbf{II}) \citep{glielmo2022ranking} as a metric to compare the representations of all models with respect to each subset. II is a non-symmetric measure of similarity that relies on the number of shared neighbors for each datapoint of two spaces A and B, by quantifying how much information is shared between the two as the similarity between local neighborhoods. Since it relies on local distances, II is robust with respect to representational spaces lying on high-dimensional manifolds \citep{ acevedo2026quantitativeanalysissemanticinformation}. II is estimated as:

\begin{equation}
2/N^2 \sum_{i,j} r_{ij}^A=1 r_{ij}^B
\end{equation}
where \(A\) and \(B\) are the two spaces, \(r^A\) and \(r^B\) are the neighborhood ranks of one point for \(A\) and one point for \(B\), \(N\) is the total number of datapoints. If II is closer to \(1\), the two spaces are not mutually predictable; if it is around \(0.5\), they are mutually and symmetrically predictable; if it is closer to \(0\), one space predicts the other. II is an asymmetric measure of predictability when it is closer to \(1\) or \(0\), but since our results were mostly symmetric with II values around \(0.5\), we adopted a symmetrized value of II, by averaging the two resulting values from any analysis \citep{acevedo2026differential}.
For this study we used an adaptation of the implementation of II from the
public repository \href{https://github.com/sissa-data-science/DADApy}{DADApy}~\citep{dadapy}.

\section{Results}
We extracted the features of all the datasets (see Figure~\ref{fig:texture_spectrum}) and from all the models. For VGG-19 we extracted the batch-normalization layer following each convolutional block, consistently with previous literature (\cite{depaolis2026, gatys2015}). For CLIP, DINO-v2, and iGPT we considered the average of each layer's final hidden state as a global representation of the token sequence; in the case of CLIP and DINO-v2 we excluded the end-of-sequence CLS tokens. Since ViT activations show the presence of massive outliers connected to the attention sinking effect \citep{son2024prefixing, kaul2025attention, kang2025visual}, we clipped the activations of CLIP, DINO-v2, and iGPT to retain the .95 percentile \citep{Ma2019, kim2021self}. This passage was unnecessary for VGG-19 activations, as they tend to be more stable.

We then performed the following analyses: (i) a comparative analysis of the models' representation for each subset, to assess whether the models generate similar or dissimilar representations for the same type of image; (ii) a comparative analysis across the representations of the image types generated from the same DTD source  -- therefore Noise, V\&C, P\&S, G, plus DTD -- for each model, to assess whether each model can form representations that display meaningful relationships among the images; (iii) we compared the models' response to the previous analysis against human performance collected in the psychophysics experiment, to test whether the representations across textures found in the ViTs align to human perception better than those of the CNN; (iv) the study of the effect of different images on the representational geometry of all models
(see Appendix ~\ref{model_geometry}).

\begin{figure}[h]
\centering
\includegraphics[width=13.5cm]{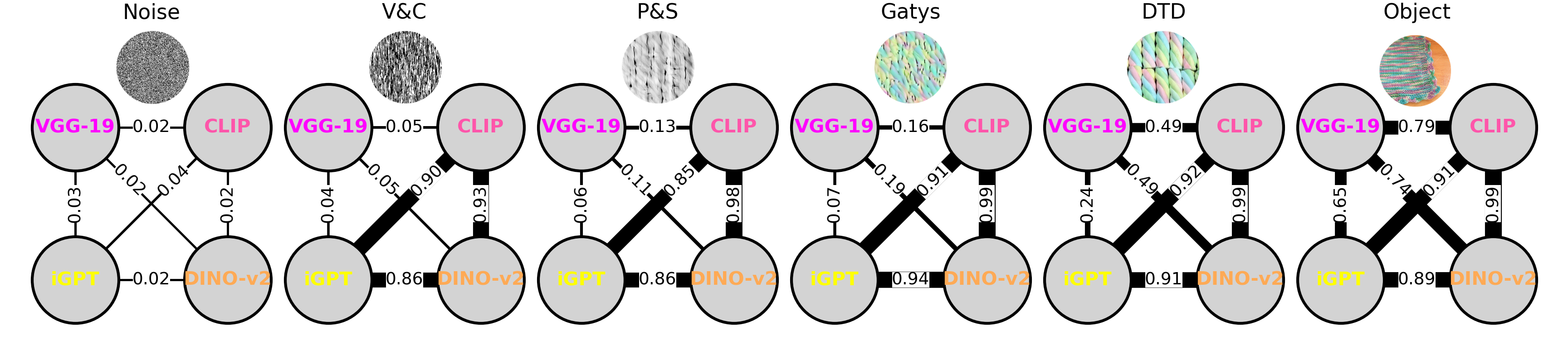} \\
\caption{Graphs representing the symmetrized mean II across models. For visualization purposes, for each subset we selected the best II between the two most mutually predictable layers for all models. For ease of illustration, we used \(1-II\): strong links and high values show high mutual predictability; thin links and low values represent low predictability. An extended figure is available in Appendix ~\ref{across_models_appdx}, Figure~\ref{fig:across_models_ii}.}
\label{fig:across_models_graph_2}
\end{figure}

\paragraph{ViTs show agreement in representing the same image type.} Here we compare the models’ representations for each type of input image. For each dataset, we compute II among all the possible pairs of models. In figure~\ref{fig:across_models_graph_2}, we show a graph illustration of the results of this analysis: for each subset, one graph has four nodes, one per model, and six links, indicating the \(1-II\) value computed for each pair of models. The links are unidirectional because we symmetrized II, and for illustrative purposes, we converted II into \(1-II\). We focus on the layer's representations with the largest \(1-II\) values for each model pair. A full plot for all the layers can be found in Appendix ~\ref{across_models_appdx}.  

When Noise is the input dataset, we observe only thin links connecting the models, with extremely low \(1-II\) values: this happens because the representations formed by the four models are extremely different and there is no mutual predictability between them. If we observe the case of V\&C, we notice a radical change: already when representing simple textures, there are strong links among CLIP, DINO-v2, and iGPT, as well as weak links between VGG-19 and the three ViTs. The strong links mean that the representations formed by CLIP, DINO-v2, and iGPT are similar and predictable by each other, while the thin links  between VGG-19 and the three ViTs signify non-mutually predictable representations. This pattern keeps being present and evolving as the textures become semantically richer: in the cases of P\&S and G, CLIP, DINO-v2 and iGPT still hold strong links among each other, while links with VGG-19 are weak. When DTD is the input dataset, the situation changes: links among ViTs are still strong, but the links to VGG-19 become stronger, with \(1-II \simeq 0.5\). This means that the representations between VGG-19 and the ViTs become mutually predictable. In the case of Object, we still see strong links among ViTs, and increasingly stronger links involving VGG-19, with \(1-II \simeq 0.75\): in particular, the Object graph shows that the representations of complex and semantically rich images are largely shared and aligned across all models, regardlessly of their architecture.

These results suggest that VGG-19 and ViTs rely on different representational strategies. While ViTs capture the representations of the same textures as early as in V\&C, and maintain this representational pattern stable through the complexity evolution of textures until DTD, VGG-19 instead seems to form representations of textures that differ according to the complexity: semantically poor textures elicit representations that do not align well to those of the ViTs, while semantically rich textures representations and objects representations largely align to those of the ViTs.

\begin{figure}[ht]
\begin{centering}
\includegraphics[width=14cm]{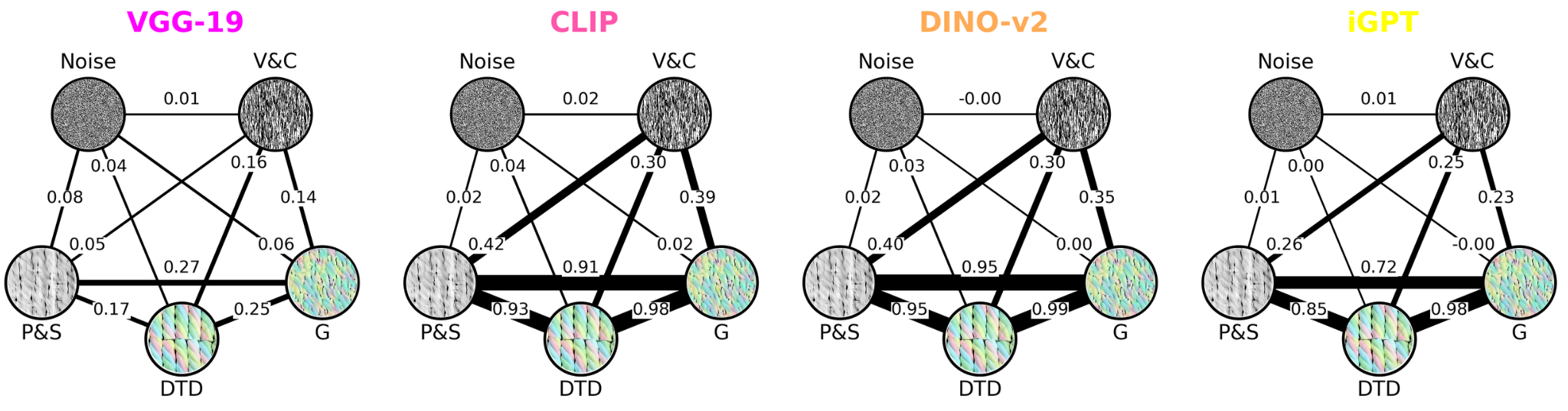} \\
\caption{Graphs representing the symmetrized mean II across datasets, excluding Object. For representational purposes, for each model we selected the II between the best mutually predictive layers of all the subset combinations. For illustrative purposes, we inverted the II scale to \(1- II\): strong links correspond to high mutual predictability; thin links represent low predictability; slightly thick links represent equal mutual predictability. An extended figure is available in Appendix ~\ref{across_data_appdx}, Figure~\ref{fig:across_data_ii}.}
\label{fig:across_datasets_graph}
\end{centering}
\end{figure}

\paragraph{ViTs capture relationships among textures of different types.}\label{across_data} In this analysis, we investigate the relationship between different textures in each model's representation. For each model, we compute II between all the possible combinations of textures: we consider Noise, V\&C, P\&S, G, and DTD. Once again we show a graph representation of the results of this analysis in figure ~\ref{fig:across_datasets_graph}: for each model, one graph has five nodes, one for each type of input image, and ten links, one for each pair between input types. Also in this case the links are unidirectional because we symmetrized II values. For illustrative purposes, we converted II to \(1-II\), and we visualized only the model's layer representation with the best II values. A full plot of the analysis with all the layers can be found in Appendix ~\ref{across_data_appdx}.

The graph of VGG-19 shows weak links connecting any data type considered, Noise, V\&C, P\&S, G, DTD, where the best \(1-II\) is around \(\sim 0.1\). This means that the representations of VGG-19 of any image type are unrelated to each other. When we consider CLIP instead, we observe strong links and \((1-II) \geq 0.9\) among P\&S, DTD, and G textures, as well as medium sized links with \((1-II)\sim 0.35\) with respect to V\&C textures. We instead observe thin links with \((1-II)\sim0\) between Noise and the four types of textures. This result shows that CLIP representations of P\&S, DTD and G are very similar and highly mutually predictable, V\&C representations are somewhat mutually predictable to the other three textures, and Noise representations are not mutually predictable with respect to any texture. Both DINO-v2 and iGPT show the same pattern as CLIP: strong links among P\&S, DTD, and G, medium-strenght links between V\&C and the other three textures, extremely weak links between Noise and all the textures.

These results suggest that while CLIP, DINO-v2, and iGPT form largely overlapping representations between high and middle-level textures as in P\&S, DTD, G, and form quite similar representations to V\&C, VGG-19, instead, does not produce shared or overlapping representations among any image type. The absence of any representational similarity with respect to Noise in any of the three models is not surprising, since it belongs to a different distribution other than that of the DTD source. In this analysis, we excluded Object because a point-wise comparison with the other subset using II is unfeasible.

\paragraph{ViTs align with human perception in representing textures.}
Finally, we compare the similarity across datasets according to the various models, discussed in the previous analysis, against human performance, to test which model best aligns with human perception of textures.

\begin{figure}[h]
\centering
\begin{minipage}{0.3\textwidth}
    \centering
    \includegraphics[width=\textwidth]{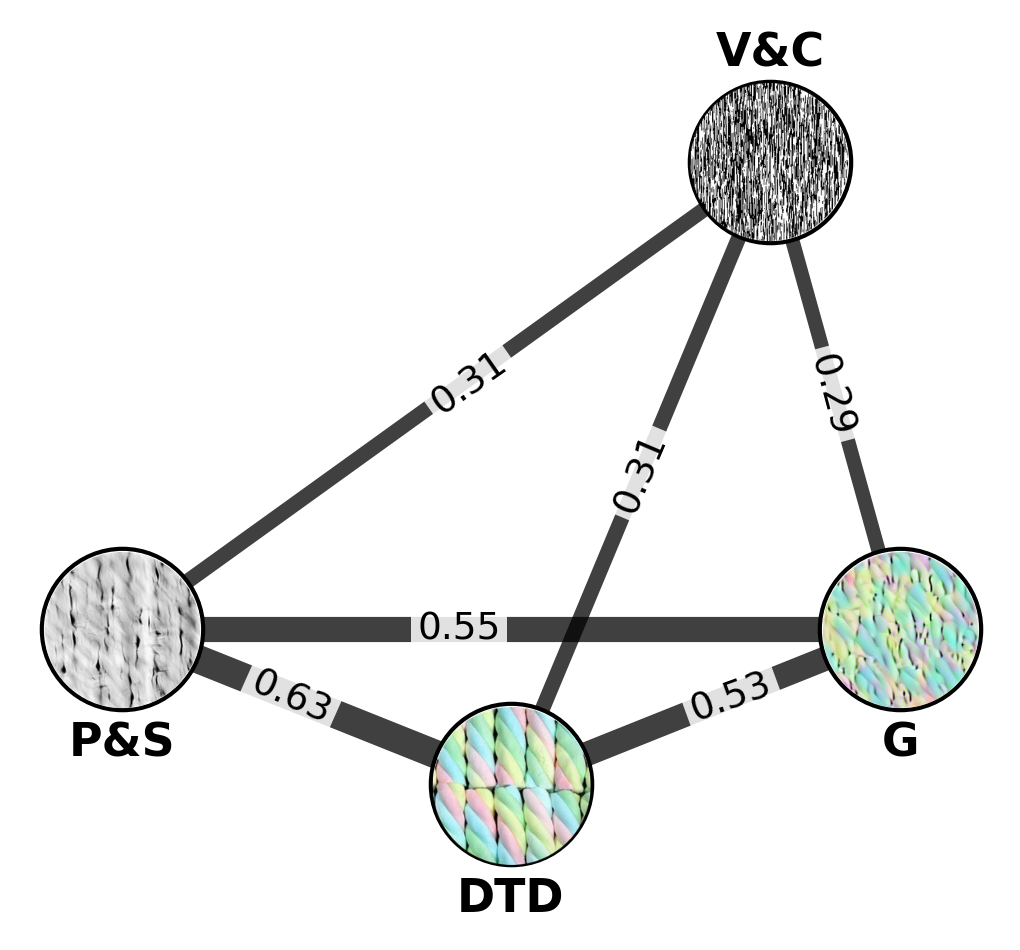}
    \subcaption{Human accuracy graph.}
    \label{fig:human_accuracy_graph}
\end{minipage}
\begin{minipage}[c]{0.6\textwidth}
    \centering
    \includegraphics[width=\textwidth]{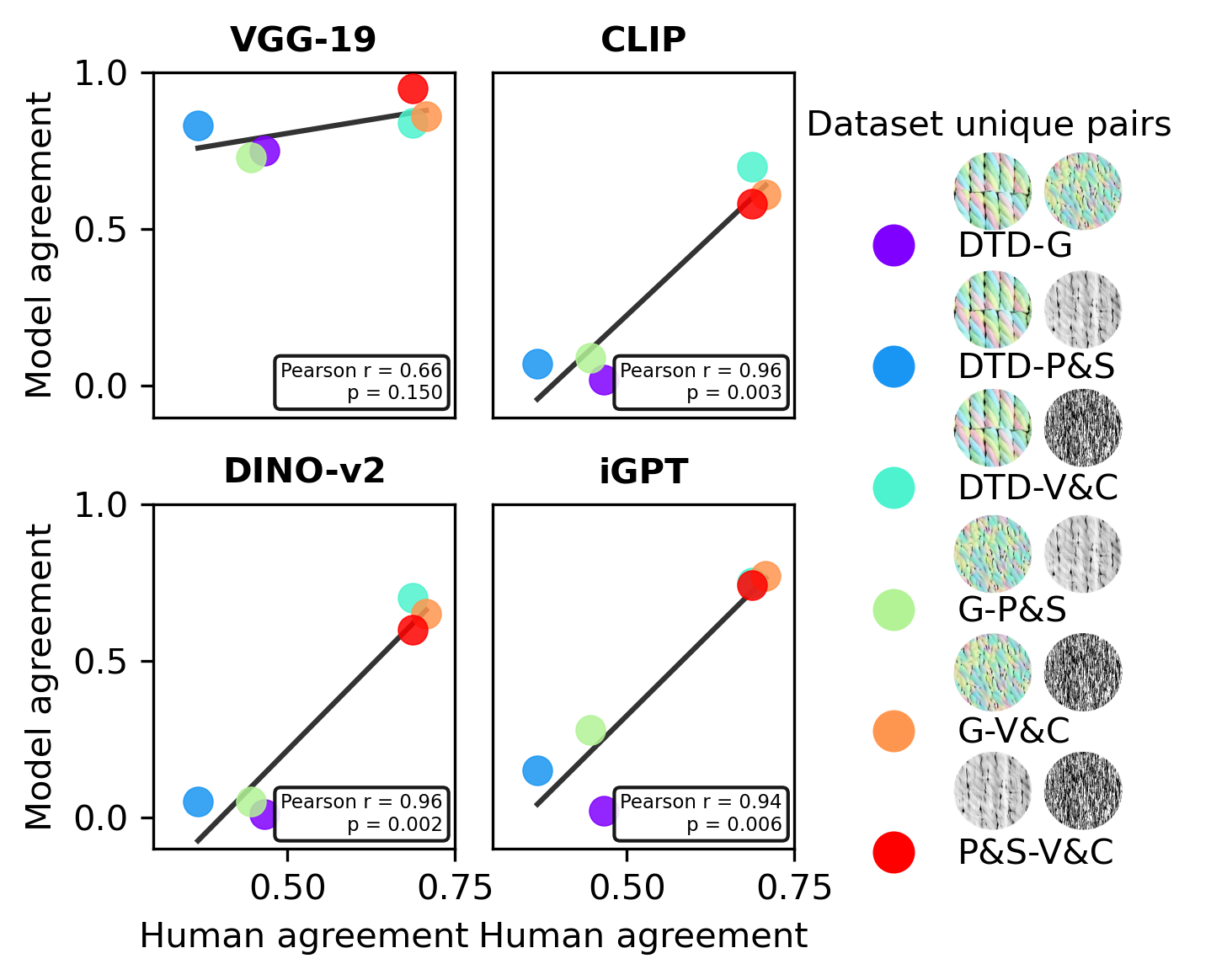}
    \subcaption{Correlation between human accuracy and model II scores.}
    \label{fig:human_models_correlation}
\end{minipage}

\caption{
Human pairwise accuracy and its correlations with models performance. For illustrative purposes, in panel~(a) we represent $1-\mathrm{accuracy}$, so that thicker links mean stronger perceptual overlap. Panel (b) shows the four correlation plots among humans and the four models. We used Pearson's r and we report \textit{r, p} inside the boxes. The legend maps a color for the II score used in the correlation for each pair of image types.
}
\label{fig:human_accuracy_summary}
\end{figure}

We obtained one human accuracy score for each pair of texture subsets by averaging accuracy across participants, and we represented accuracy values as undirected links in a graph. Finally, we run a correlation test with Pearson's $r$ to assess which model best matches human perception. For visualization purposes, figure~\ref{fig:human_accuracy_graph} shows a graph representation of human \(1-accuracy\) scores as links among V\&C, P\&S, G, and DTD textures, represented as nodes. We observe thick links and medium-high \(1-accuracy\) scores among P\&S, DTD, and G, plus medium-strength links V\&C and the three other textures. This result shows that participants easily distinguish V\&C textures from the rest of the textures, when V\&C is presented as the odd; on the other hand, humans find more difficult to tell apart P\&S, DTD, and G when either of them is presented as an odd. This indicates that high-to-medium level complexity textures are perceived as similar to each other, while semantically poor textures are perceived as outliers, that are thus easier to tell apart. Figure~\ref{fig:human_models_correlation} shows the result of the correlation analysis. The colored dots are II values found in the analysis in paragraph~\ref{across_data}, coded in the legend for each pair of textures, that are correlated to the participants' accuracy scores. We observe that, while VGG-19 does not show any significant correlation with human performance, CLIP, DINO-v2 and iGPT all show high and significant correlation values.

These results suggest that ViTs representations of textures that display different levels of complexity align better to human texture perception than those of VGG-19.

\section{Discussion}
In this study we investigated the representation of visual textures in artificial neural networks and the alignment between deep neural network representations and human perception of these textures. We built a large dataset of visual textures characterized by different degrees of perceptual complexity. We considered two existing databases: ImageNet for object images, and Describable Textures Dataset for naturalistic textures; we then used three texture synthesis algorithms to generate textures with progressively simpler content starting from DTD: Gatys' algorithm, Portilla \& Simoncelli's algorithm, Victor \& Conte's model. We also considered a set of randomly generated images.

We explored four models: VGG-19, CLIP, DINO-v2 and iGPT. VGG-19 is a CNN and a popular model of the mammalian ventral visual stream; CLIP, DINO-v2, and iGPT are ViTs, and their alignment to human vision is largely unexplored. We extracted features from all models with all the image types: Noise, V\&C, P\&S, G, DTD, Object. We collected human accuracy scores with a odd-one-out texture discrimination task and analyzed the relationship among the models' representations in several ways with Information Imbalance, as follows: 
(i) to quantify the agreement between the representations of the same image type for all models; (ii) to quantify the relationship captured by the models' representations for different textures; (iii) to compare human perception against the models' performance.


We find that ViTs produce overlapping representation of all textures and Object, while VGG-19's representations are different and become similar to those of ViTs only when DTD and Object serve as input. We therefore conclude that the architecture may be the main factor explaining the difference in performance between the ViTs and the CNN as it an important shared element among the three ViTs.

We furthermore observe that while the three ViTs can form largely overlapping representations of P\&S, DTD, and G, and somewhat similar representations between those textures and V\&C, VGG-19 doesn't form any similar representation among the four types of textures. Once again, we connect this result to the architecture: ViTs are equipped with attention, designed to build representations with both local and global range correlations among tokens \citep{vaswani2017, chen2020igpt}. Convolutions and pooling instead generate representations that rely only on spatial-bound and local image features, although effective spatial ranges grow with layer depth. We conclude that attention plays a role in the formation of image representations that are stable and shared across models with the same architecture but different objectives, consistently with the Platonic Hypothesis \citep{pmlr-v235-huh24a}, which are robust to the progressive perceptual loss of the stimuli. 

Finally, we find that human perception of texture is better aligned with the ViT representations than with the CNN representations. This result suggests that ViTs, despite not being commonly recognized as a good model of human vision, outperform VGG-19 in at least certain perceptual dimensions.


Taken together, our results suggest that ViTs might be useful models of human texture perception, and should be considered as valid models to expand research about human visual perception, perceptual organization and biological modeling.
More broadly, our findings underscore the idea that texture and object processing are only partially intertwined. Building robust models of texture perception may therefore require moving beyond systems trained exclusively for object recognition, while still accounting for the contribution of high-level semantic information. Our results also point to a potential limitation in the way Julesz's conjecture has been operationalized in computational models of texture perception. Rather than relying exclusively on local image feature correlations, our findings suggest that such models may need to adopt a broader representational perspective than that originally implied by Julesz's conjecture.

\section{Limitations}
Due to limited computational resources, we could not include other CNN models in the analyses. If this could be done, based on the representational similarities for textures across different CNN architectures~\citep{depaolis2026}, we would expect them to show similar results as VGG-19 and to be dissimilar to ViTs.


\bibliographystyle{plainnat}
\bibliography{bib}

\section{Appendix}

\subsection{Supplementary analysis}

\subsubsection{Effect of texture type on each model's signature geometry.} \label{model_geometry}
\begin{figure}[ht]
\begin{centering}
\includegraphics[width=13cm]{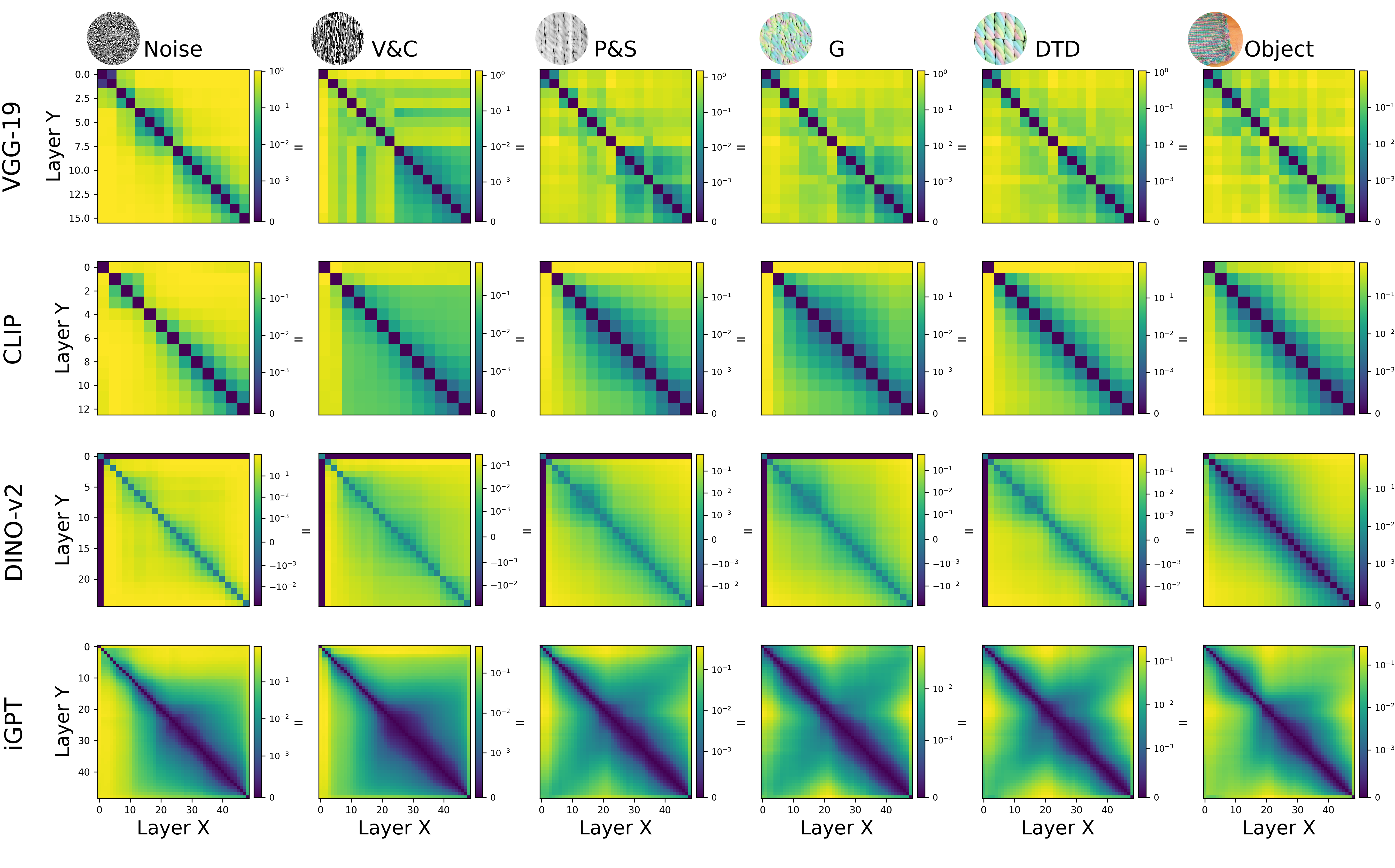} \\
\caption{Heatmaps for each model (rows) by each dataset (columns). For each matrix, the entries are the symmetrized mean II values calculated between all the possible combinations of layers. Each matrix is equipped with its relative color-bar of II values in logarithmic scale.}
\label{fig:within_mean_ii}
\end{centering}
\end{figure}

We focus on the effect of the type of input image on the representational geometry pattern of each model. For each model and each data type, we compute the symmetrized II between all the pair combinations of layers' representations. The matrices in figure ~\ref{fig:within_mean_ii} show the representational geometry of the models induced by the data type. Each matrix color coding is relative to the specific data-per-model combination and own II values, represented in logarithmic scale and reported in the legend at the side. The relative maximum II value is colored in yellow, while the minimum value, always at \(0\), is colored in dark blue. 

The first row reports the internal representations of VGG-19. If we observe the matrix for VGG-19/Objects, the entries are the II values for each possible combinations of layers (for details about the number of layers see Appendix Table ~\ref{table:models_details}). In this case, the input stimuli elicit a representational pattern featuring two dark clusters, one at the top of the diagonal and the other at the bottom. The information shared among close layers is relatively high because their representations have II values close to \(\sim0\). The color of the heatmap goes progressively towards yellow as we move away from the diagonal, because II values are around the maximum. In this case, the pairwise layer representations do not share much information. This happens for layer pairs that are distant from each other. In the case VGG-19/DTD, we observe a similar color pattern to that of VGG-19/Objects: two darker clusters, one at the top of the diagonal and one at the bottom, where representation of close layers are similar; when moving away from the diagonal yellow prevails because the representations between distant layers become dissimilar. This representational pattern seems to fade in VGG-19/G and VGG-19/P\&S, and it gets completely disrupted in VGG-19/V\&C: here, in fact, we observe a large dark block at the bottom half of the diagonal, where representations between the last group of layers are similar, but also redundant because they don't display further information shifts. The representations between the other layers' pair combinations feature lighter colors, signaled by intermediate II values due to a certain degree of similarity. 

Moving on to the second row and observing the heatmap of CLIP/Object, we notice two main dark blocks, one located at the top half of the diagonal and the other at the bottom half. A dark coloration is present around the diagonal, which becomes progressively lighter as the entries are II values calculated between distant layers' representations. This pattern is consistent to what we observed in VGG-19/Objects: Objects elicit representations that tend to be similar between closer layers, with II values around \(\sim 0\), and dissimilar between layers located farther away, with II values towards the maximum. In the case of CLIP/DTD, we observe a similar color pattern, with less defined clusters, but still featuring lower II values around the diagonal. The pattern starts to break from DTD/G and across DTD/P\&S, where we observe an uniform dark coloration stemming from the diagonal, which only gets lighter when layer n.\(1\) is compared against all the other layers. This means that CLIP representations of respectively G and P\&S textures are similar across the whole model, with a single information shift between the input layer and the rest of the layers. The pattern completely disappears in CLIP/V\&C: the information shared between layers n.\(1\) and n.\(2\) is quite low, while it is redundantly high in the representations between rest of the layers' combination. The representational patterns observed in CLIP show a similar progression to those of VGG-19, but while the difference in information through layers in CLIP appears to evolve smoothly, in VGG-19 we observe occasional breaks signaled by lighter stripes: we connect it to the effect of the pooling on the beginning of each of the \(5\) convolutional blocks.

DINO-v2's representations follow a similar evolution to that observed in CLIP: DINO-v2/Objects shows dark blocks around the diagonal, with a large cluster in the upper half of the diagonal and a second cluster in the bottom half. The blue central part, featuring low II values among close layers, fades to yellow when II is computed among distant layers. We find the same pattern in DINO-v2/DTD, with less defined clusters around the diagonal. This characteristic representational geometry, similarly to CLIP, gradually dissipates as the textures become simpler, through G, P\&S, and V\&C: in these cases, the clusters become less and less structured, resulting in a general darkening of the whole heatmap, except dark stripes in the input layer where II is \(\sim0\). 

In the row of iGPT, we find a slightly different signature geometry, probably due to the following characteristics: it organizes its representations of images as an  encoder-decoder even if the architecture backbone is an decoder only \citep{chen2020igpt, Valeriani2023, Mahaut2026}; it is deeper than the other architectures (see Appendix Table~\ref{table:models_details}); the objective is next-pixel prediction. In iGPT/Object we observe an almost uniform medium-dark coloration through all the layers, with four yellow spots in the middle of the four sides of the matrix: this means that the representation is generally self-similar among all layers, however they become most dissimilar when II is computed between intermediate layers and early layers, and between intermediate layers and late layers, respectively. In practice, iGPT representational shift happens around the bottleneck at the center, while in VGG-19, CLIP, and DINO-v2 the representational bottleneck is around the output layer, as we show with more classical linear methods (see Appendix Supplementary Figure ~\ref{fig:linear_probe_combined}). Also in the case of iGPT the signature pattern progressively dissipates as the input textures becomes simpler: in iGPT/DTD the representational geometry resembles that of iGPT/Object, with a meaningful information shift in the middle layers. Then, through iGPT/G and iGPT/P\&S, the information shift in the middle becomes gradually less confined to a few middle layers. And finally, in iGPT/V\&C, II tends to be redundantly low, the middle-layers information bottleneck disappears, switching to a single, initial information shift between early layers and the rest. 

In the case of Noise images, across VGG-19, CLIP, DINO-v2, and iGPT layers representations, we always observe a prominent yellow color given by high II values, accompanying a darker area surrounding the diagonal. The only exception is iGPT, where dark and light colors with respectively low and high II values, both occupy about half of the heatmap. In all models, the representational geometry elicited by Noise is compatible to representations that are always changing and share little information with previous layers. This is due to the randomness and unpredictability of the information carried in noisy images, which at any processing step elicit a certain degree of surprisal in the models. 

Taken together, these results suggest that the representations of Objects and those of semantically rich texture such as DTD rely on a similar strategy in all models: the representations in each group of close layers are well structured. However, as the textures gradually lose semantic content and become simpler, the representational strategy adapts to the lack of semantic content of the images and becomes more redundant with only one relevant shift: most information is elaborated in the first layers, while the rest is just repetition. Surprisingly, while the four textures types belong to the same distribution of the source database DTD, only the representations of DTD seem to resemble those of Object, even though Object is drawn from a different distribution with a different semantic labeling system. We conclude that the type of input, and in particular its semantic content, has an effect on the representational geometry of all models. Semantically rich images and textures show representations structured as clusters, while semantically poor textures produce more sparse representations. This experiments highlights that, despite differences in architecture and objectives between the CNN and the ViTs, the progressive loss of complexity in the input images has an effect on the representational organization of each model, which tends to rely on all layers in semantically-rich images, while it tends to rely only on the first layers in low-complexity textures.

\subsubsection{Linear probe.} In order to assess whether the analysis done with Information Imbalance, a highly non-linear metric, can be also performed with linear model, we build a linear classifier to classify images of textures generated from the source DTD into the original labels found in DTD.

For each model we build three linear classifiers, one for an early, one middle, and one for a deep layer. We then perform the following experiments (a) classification: for each model, we trained and tested three linear probes to classify their features into their original classes with a 80-20 train-test regime; (b) transfer learning: for each model, we trained the three probes to classify 80\% of the DTD features and tested them to classify 20\% of the features from the remaining subsets. We excluded Object subset because it does not share the classification structure with the other textures subsets. 

\begin{figure}[ht]
\centering
\begin{subfigure}{1\textwidth}
\begin{centering}
\includegraphics[width=10cm]{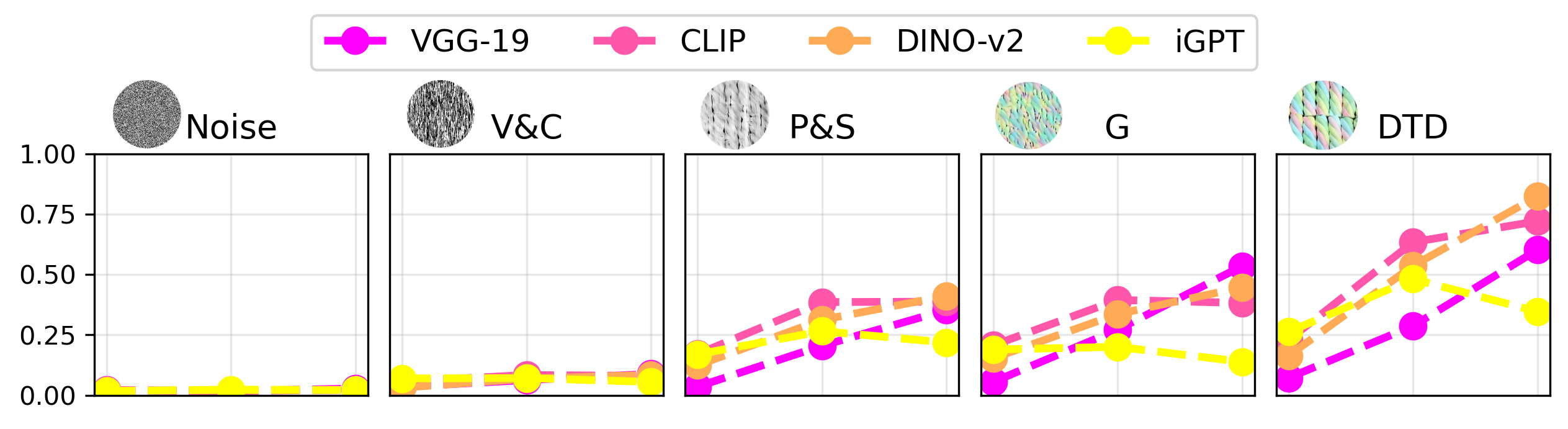}
\end{centering}\caption{Classification}
\end{subfigure}

\vspace{0.1cm}

\begin{subfigure}{1\textwidth}
    \begin{centering}
        
\includegraphics[width=10cm]{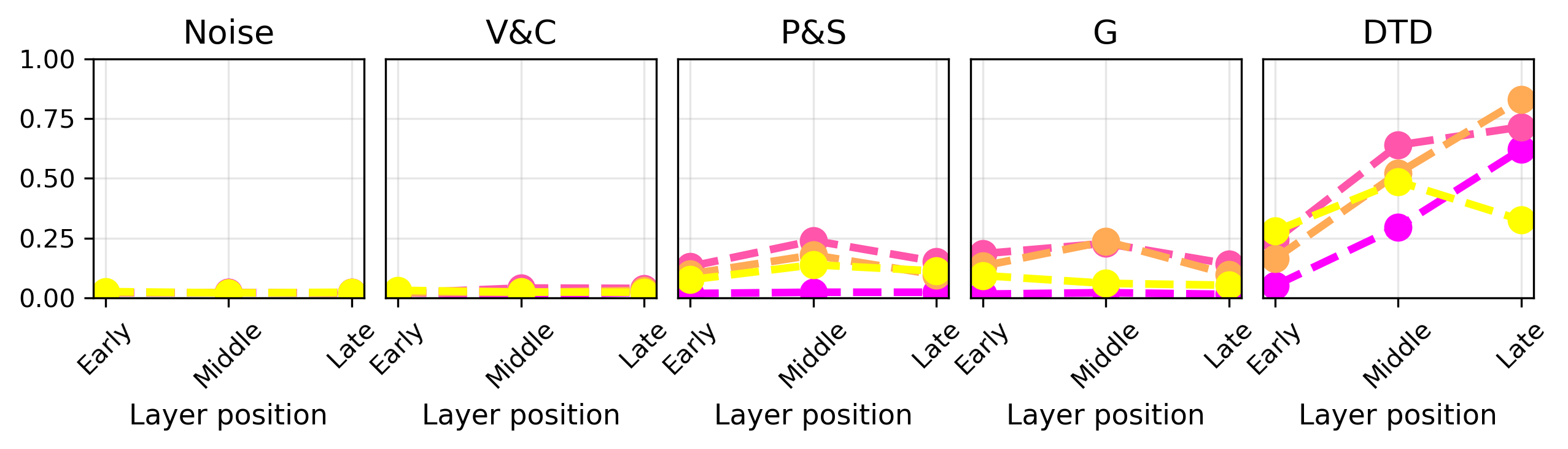}
\end{centering}
\caption{Transfer learning}
\end{subfigure}

\caption{
(a) Classification: accuracy scores for a triplet of probes trained on early, middle, and deep extracted features for each dataset.
(b) Transfer learning: accuracy scores for a triplet of probes trained on early, middle, and deep layers' extracted features of DTD and tested on the other datasets.}
\label{fig:linear_probe_combined}
\end{figure}

Figure~\ref{fig:linear_probe_combined} shows that (a) the classifiers increase in accuracy as the data become more complex. iGPT peaks accuracy in the middle layer because it is an autoencoder, while the other models peak in the last layer. This behavior agrees with previous studies \citep{Valeriani2023, Mahaut2026}; (b) the classifiers fail at generalizing features learned from DTD. We conclude that a linear classifier can't capture the relationship among subsets, while II can do so due to its nonlinearity.

\clearpage

\subsection{Supplementary figures}

\subsubsection{ViTs show agreement in representing the same image type.}\label{across_models_appdx}
\begin{figure}[h]
\begin{centering}
\includegraphics[width=15cm]{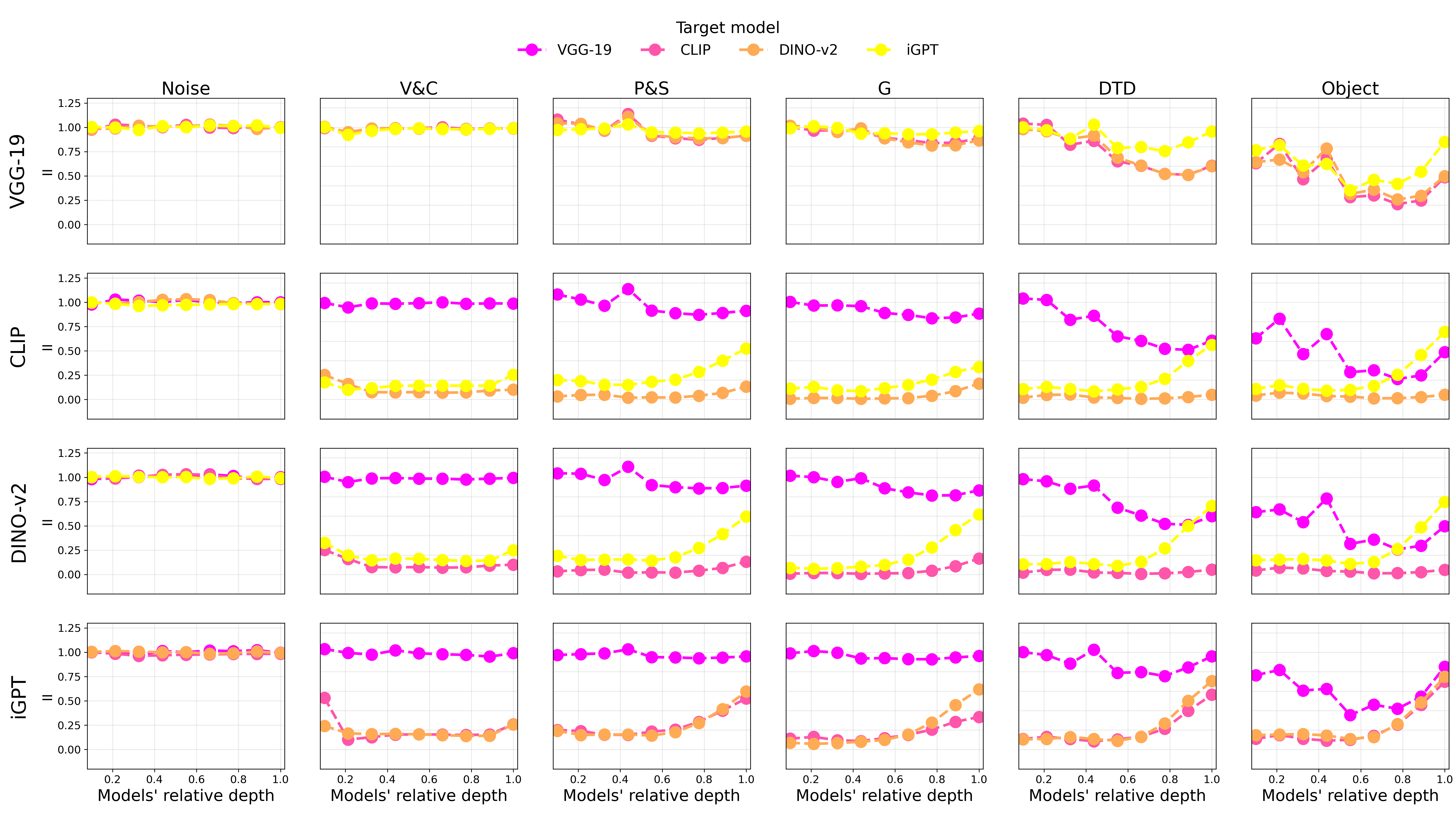} \\
\caption{Extended plot for \textbf{Section 4.2, Figure~\ref{fig:across_models_graph_2}}: For each dataset (columns) we compute II for each pair of  models. In particular, as shown in the rows, each model serves as a \textit{reference}, and the other three serve as \textit{target}, as shown in the columns and in the top legend. We quantify the depth of each feature's layer as the relative percentage, and compared symmetrized II scores between reference layers and target layers along the same relative depth.}
\label{fig:across_models_ii}
\end{centering}
\end{figure}
\clearpage

\subsubsection{ViTs capture relationship among textures of different types.}\label{across_data_appdx}
\begin{figure}[ht]
\begin{centering}
\includegraphics[width=13cm, height=8cm]{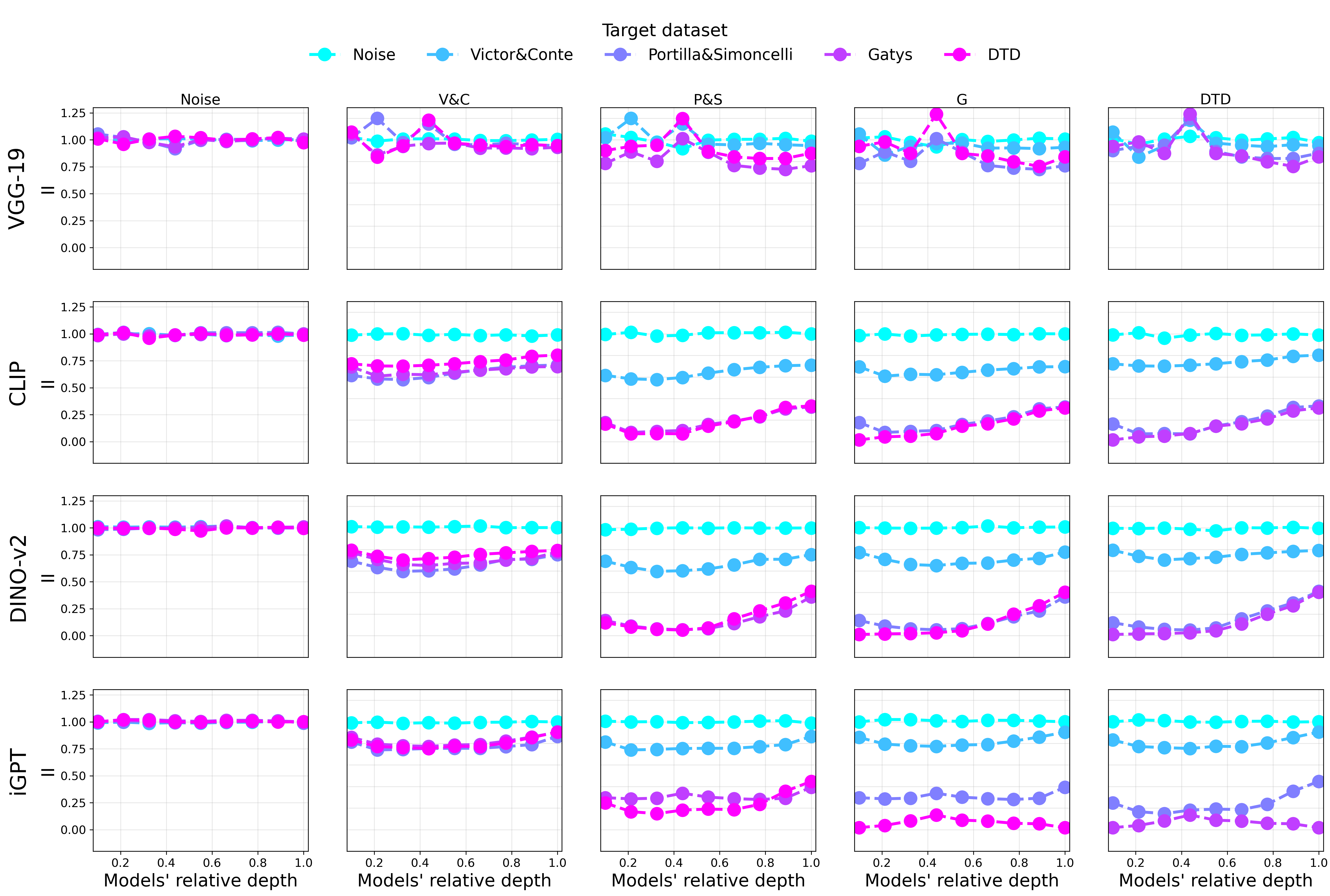} \\
\caption{Extended plot for \textbf{Section 4.3, Figure~\ref{fig:across_datasets_graph}}: For each model (rows) we compute II between all pairs combination between a \textit{reference dataset} (columns) and four remaining \textit{target datasets} (legend). We quantify the depth of each feature's layer as the relative percentage, and compare symmetrized II scores between reference layers and target layers along the same relative depth.}
\label{fig:across_data_ii}
\end{centering}
\end{figure}
\clearpage

\subsubsection{Human accuracy}
\begin{figure}[h]
\begin{centering}
\includegraphics[width=10cm, height=7cm]{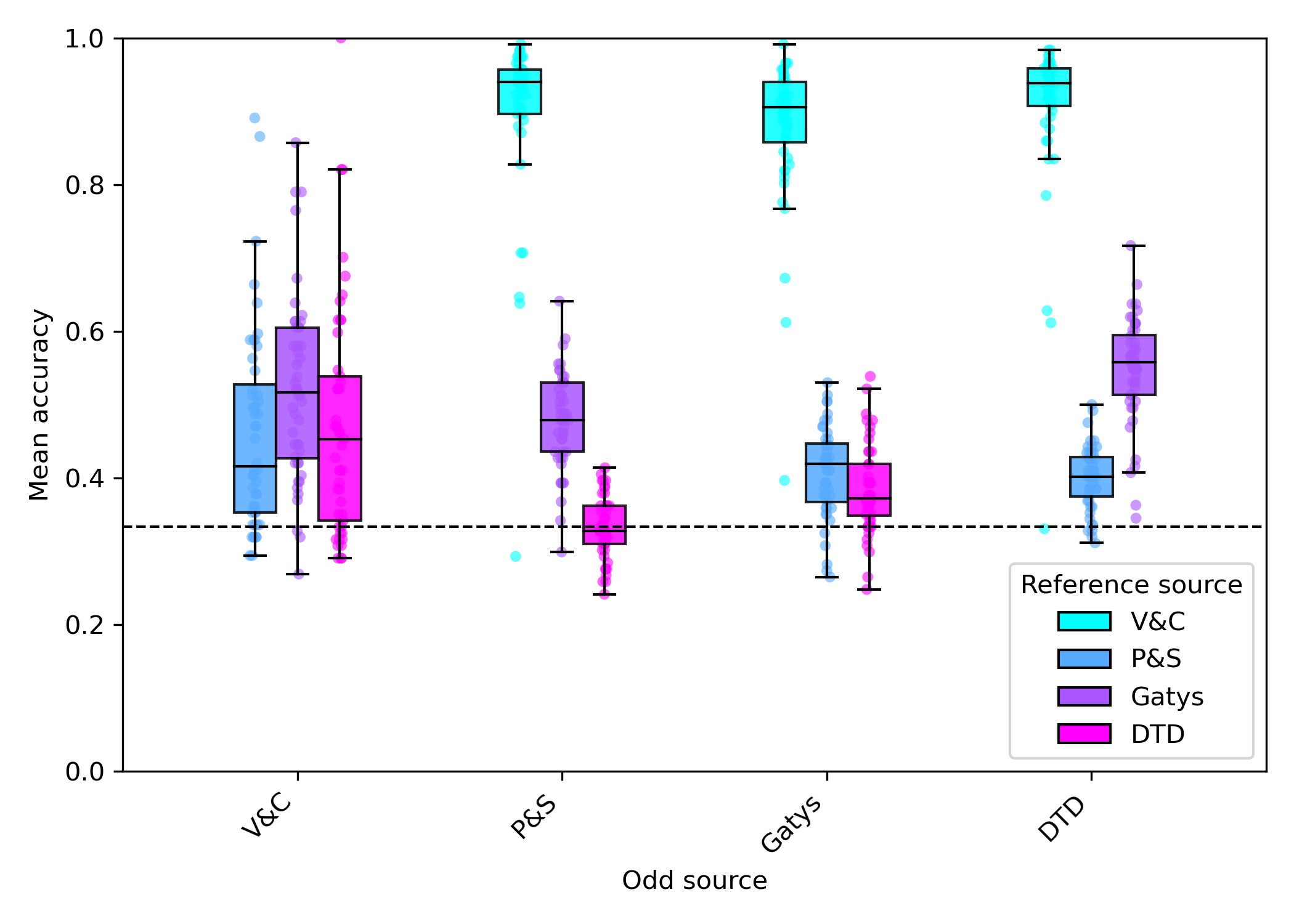} \\
\caption{Extended plot for \textbf{Section 4.5, Figure~\ref{fig:human_accuracy_graph}}:
This plot represents the relationships among texture subsets for each pair of \textit{odd image} (see X axis) and and \textit{reference image} (legend) as human accuracy scores (see Y axis) of all participants collected in the behavioral task. Each box summarizes the distribution of mean accuracies for one combination of odd source and reference source. The scatter shows the individual data points, the box shows the interquantile range, the central line shows the median, and the errorbars show the non-outlier range. We observe that accuracy is highly odd source-dependent. Trials with V\&C as the source were solved with almost top accuracy for all odd-source conditions, indicating that V\&C images form a perceptually coherent and distinct subset. In contrast, when V\&C was the odd source, performance was moderate: accuracy in distinguishing P\&S, G and DTD gets lower, suggesting a substantial perceptual overlap.
This asymmetry suggests that odd-one-out performance is not determined by a symmetric perceptual distance between datasets, but also by the internal variability of the reference source.}
\label{fig:accuracy_humans}
\end{centering}
\end{figure}

\subsection{Stimulus dataset generation}\label{stim_data_gen}
\subsubsection{Victor and Conte}
We used this algorithm to generate textures of type \textbf{V\&C}. This approach produces low-complexity binary textures that have been employed in a number of efficient coding studies \citep{hermundstad2014,caramellino2021,zanonPredisposedLearnedPreferences2026} due to their appealing analytical properties and controllability and despite being perceptually distant from natural stimuli. The algorithm allows to control explicitly all local image statistics measured on \(2\times2\) pixel blocks \citep{victor2012}. By exploiting an assumption of translational symmetry, the 16 possible block statistics are reparameterized into ten coordinates:
\begin{itemize}
    \item luminance bias: \(\{\gamma\}\)
    \item pairwise correlations: \(\{\beta_{-}, \beta_{\mid}, \beta_{\backslash}, \beta_{/}\}\)
    \item three-point correlations: \(\{\theta_{\llcorner},\theta_{\lrcorner}, \theta_{\ulcorner},\theta_{\urcorner}\}\) 
    \item four-point correlations: \(\{\alpha\}\)
\end{itemize}
Given chosen values of one or more of these coordinates, the remaining image statistics are fixed implicitly by a maximum-entropy principle: the generated texture is the most random binary image ensemble compatible with the specified local constraints. To generate the texture of type \textbf{V\&C} corresponding to a given DTD image we first computed the block statistics following a procedure analogous to that described in \cite{hermundstad2014}. Then, we determined which among the 10 parameters was most strongly biased away from 0, the point in parameter space that indicates full randomness. Finally, we sampled a random texture constrained to have that same value of the parameter using METex \citep{PiasiniMetex}.

\subsubsection{Portilla and Simoncelli}
We used this algorithm to generate textures of type \textbf{P\&S}. It works particularly well on regular and homogeneous textures, returning spatially regular and gray-scaled synthetic images of textures. Given an exemplar texture, the image is
first decomposed using a steerable pyramid. The algorithm then measures a fixed set of statistics on the resulting wavelet coefficients. A synthetic texture is generated by starting from a random image sampled from \(\hat{x} \sim \mathcal{N}(0,1)\), and iteratively projecting the image onto the constraint sets defined by the target statistics, until the synthetic image matches
the exemplar statistics.

We used the algorithm with the following parameters (see \cite{portilla2000} and \cite{portilla_simoncelli_texturesynth} for reference implementation of the algorithm):
\begin{itemize}
	\item \texttt{-n}: depth of steerable pyramid = 5
	\item \texttt{-k}: number   of orientations of steerable pyramid = 4
	\item \texttt{-n}: pixel distance for calculation of auto-correlations = 7
	\item \texttt{-iter}: number of iterations = 10
\end{itemize}

\subsubsection{Gatys}
We used this algorithm to generate textures of type \textbf{G}. Type G Textures are characterized by strong perceptual and semantic similarity with the original textures, resulting in  high quality RGB images.
The algorithm's backbone is VGG-19\_bn, and the texture representations are extracted from 5 batch normalization layers compiled in to Gram matrices. For each of the 5 blocks in VGG-19 we extracted the first batch normalization layers after the first convolution of the block, resulting in 5 layers. Given an image from DTD, \(x\), the algorithm feeds \(x\) as input to VGG-19. For each extracted layer \(l\), the feature activations \(F^l\) are used to compute a Gram matrix \(G^l\), defined as:
\begin{equation}
G^{l}_{ij} = \sum_{m} F^{l}_{im} F^{l}_{jm},
\end{equation}
where the index $i$ and $j$ are the indices of the feature maps, and the index $m$ runs over all units within a feature map.
This Gram matrix summarizes the correlations between feature channels, discarding spatial information. Texture synthesis is performed by iteratively updating the pixels of an initially random image \(\hat{x}\), where pixel values are drawn from \(\hat{x} \sim \mathcal{N}(0,1)\), such that its Gram matrices match those of the original image \(x\). This is achieved by minimizing the \emph{Gram loss}, defined as a weighted sum of $L_2$ distances between the Gram matrices of the individual layers:
\begin{equation}
\mathcal{L}(\hat{x}, x) = \sum_{l} \frac{1}{4M_lN_l^2} \, \lVert \hat{G}^{l} - G^{l} \rVert^2_2.
\end{equation}
where the weighting factor is chosen to balance the contribution to the loss of each layer \citep{depaolis2026}. To minimize the Gram loss we used Adam at 30000 optimization steps. This choice allowed us to control the quality of the generated texture, by preserving some of the semantic and perceptual content of the original textures in DTD without resulting too similar.

\subsection{Models details}\label{mod_det}

\begin{table}[h]
\begin{tabular}{|c|c|c|c|c|}
\hline
\textbf{Model} & \textbf{Params count} & \textbf{Architecture} & \textbf{Objective (ImageNet)} & \textbf{N. extracted layers}\\
\hline 
VGG-19\_bn & 148M & CNN & Classification & 16 \\
\hline
CLIP-ViT-B-32 & 302M & ViT & Image-caption pairing & 12 \\
\hline
DINO-v2 & 304M & ViT & Feature extraction & 24\\
\hline
iGPT & 1,413M & ViT & Next token prediction & 49 \\
\hline
\end{tabular}
\vspace{0.5cm}
\caption{\textbf{Summary of model's details.} Consistently with previous studies, we used VGG-19\_bn. Indeed, batch normalization layers allow a stable feature representations of textures \citep{gatys2015, depaolis2026}. We extracted every batch normalization layer after any convolutional layer, for a total of 16 layers.}
\label{table:models_details}
\end{table}
\clearpage

\subsection{Human behavioral experiment}\label{hum_beh_exp}
\subsubsection{Experimental design}\label{exp_design}
We recruited 51 participants on the platform Prolific, paid £11/hour. We did not apply any particular screening with respect to age, spoken language, country of residence. Participants were required to have a normal or corrected-to-normal vision and no past history of epilepsy. The stimuli were taken from our curated datasets, cropped in circles and converted to grayscale \cite{wallis2017}. The experiment consists in 47 blocks, one for each semantic class, composed of 30
trials, for a total of 1410 trials and a median time of ~45 minutes. Images in each stimulation belong to the same semantic class. Participants were instructed to press these keys to indicate the odd: \(\rightarrow\) for right image; \(\leftarrow\) for left image; \(\downarrow\) for bottom image.

\begin{figure}[ht]
\begin{centering}
\includegraphics[width=7cm]{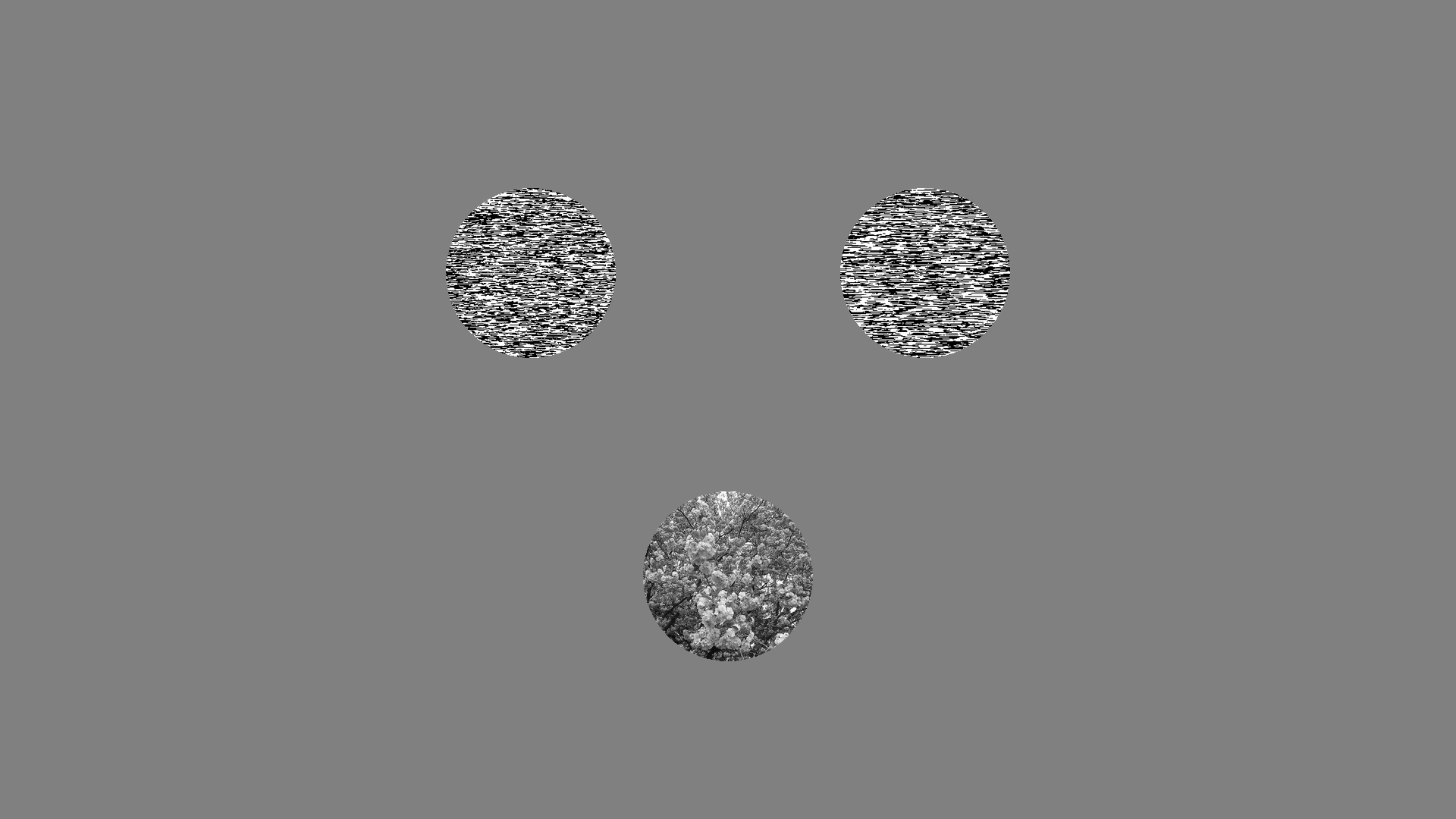} \\
\caption{Example of the visual stimulus screen in the experimental design. The odd in this case is the bottom stimulus (\(\downarrow\)).}
\label{fig:stimuli}
\end{centering}
\end{figure}

\begin{figure}[ht]
\begin{centering}
\includegraphics[width=10cm]{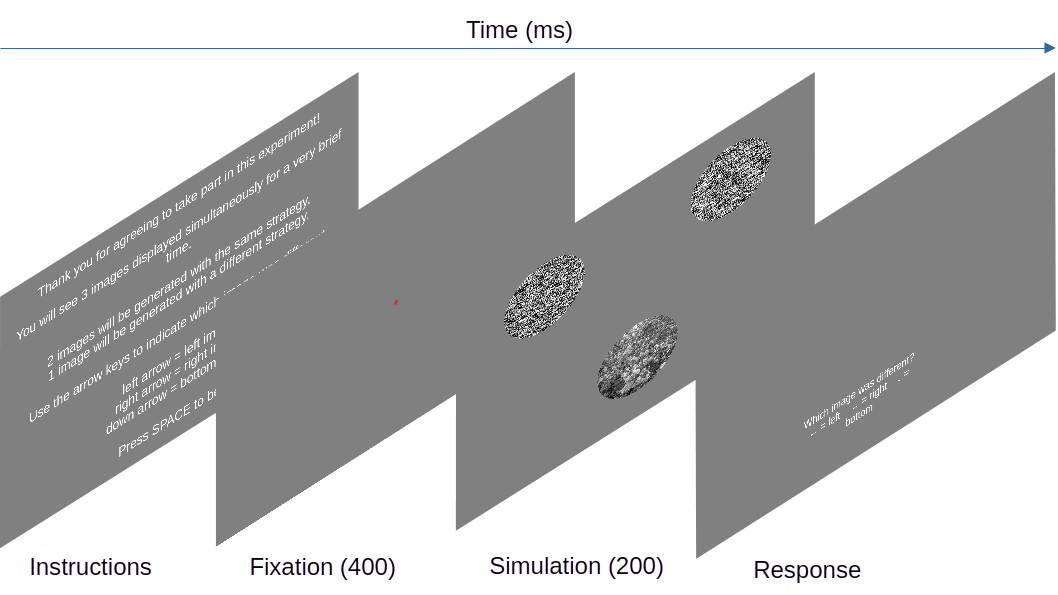} \\
\caption{Illustration of the experimental flow, with time in ms. Each simulation is paced at 200ms, proceeded by a 400ms red fixation cross. Response time is self paced. Breaks between trials are
provided.}
\label{fig:flow}
\end{centering}
\end{figure}

\subsubsection{Humans-models correlations scores}\label{exp_corr}
\begin{table}[h]
\centering
\begin{tabular}{|c|c|c|}
\hline
\textbf{Model} & \textbf{Pearson's $r$} & \textbf{p-value} \\
\hline 
VGG-19\_bn & 0.66 & 0.150 \\
\hline
CLIP & 0.96 & 0.003 \\
\hline
DINO-v2 & 0.96 & 0.002 \\
\hline
iGPT & 0.94 & 0.006 \\
\hline
\end{tabular}
\vspace{0.5cm}
\caption{Summary of table humans-models correlations for \textbf{Section 4.5, Figure~\ref{fig:human_models_correlation}}.}
\label{tab:correlations_table}
\end{table}

\section{Acknowledgments}
We thank Santiago Acevedo, Iuri Macocco and Matéo Mahaut for the useful discussions and feedback about II; Fabio Anselmi for the valuable theoretical advice; Spiros Chavlis and Dongyan Lin for supporting the initial stages of the project. EP was partially supported by the European Union –- NextGenerationEU –- PNRRM4C2-I.1.1, in the framework of PRIN Project no. 2022XE8X9E, CUP:G53D23004590001.  MB received financial support from the European Research Council (ERC) under the European Union’s Horizon 2020 research and innovation programme (grant agreement No. 101019291). AL acknowledges financial support by the region Friuli Venezia Giulia (project F53C22001770002). This paper reflects the authors’ view only, and the funding agencies are not responsible for any use that may be made of the information it contains.

\end{document}